\documentclass[11pt]{article}
\usepackage{coling2020}
\usepackage{times}
\usepackage{url}
\usepackage{latexsym}

\usepackage{xcolor}
\usepackage{dirtytalk}
\usepackage{graphicx}
\usepackage[title]{appendix}

\colingfinalcopy 

\title{Inflating Topic Relevance with Ideology: A Case Study of Political Ideology Bias in Social Topic Detection Models}
  
\author{Meiqi Guo$^1$, Rebecca Hwa$^1$, Yu-Ru Lin$^2$, Wen-Ting Chung$^3$ \\
       $^1$Department of Computer Science, University of Pittsburgh \\ 
       $^2$Department of Informatics and Networked Systems, University of Pittsburgh \\ 
       $^3$Department of Psychology in Education, University of Pittsburgh \\
       Pittsburgh, PA 15260, USA \\
       {\tt meiqi.guo@pitt.edu, hwa@cs.pitt.edu, \{yurulin, wtchung\}@pitt.edu} \\}
       
\date{}
\begin{document}
\maketitle

\begin{abstract}
We investigate the impact of political ideology biases in training data. Through a set of comparison studies, we examine the propagation of biases in several widely-used NLP models and its effect on the overall retrieval accuracy. Our work highlights the susceptibility of large, complex models to 
propagating the biases from human-selected input, which may lead to a deterioration of retrieval accuracy, and the importance of controlling for these biases.  
Finally, as a way to mitigate the bias, we propose to learn a text representation that is invariant to political ideology while still judging topic relevance. 
\end{abstract}

\section{Introduction}
\blfootnote{
    %
     \hspace{-0.65cm}  
     This work is licensed under a Creative Commons 
     Attribution 4.0 International License.
     License details:
     \url{http://creativecommons.org/licenses/by/4.0/}.
}

Due to the extensive reaches of its network and the breadth of information enmeshed in it, social media has become an invaluable data source for empirical studies in the social sciences. Yet, identifying all and only relevant information out of a vast data stream remains an untamed challenge. While topic detection methods may help researchers extract some relevant text about a topic of interest (\emph{e.g.}, immigration policies), they may miss other equally relevant text while including some irrelevant ones. Crucially, because most topic detection methods are trained, they may unintentionally contain or propagate certain biases (\emph{e.g.}, extracting more instances written by women where gender balance is expected), resulting in a skewed data collection that may lead social scientists to draw incorrect conclusions. This paper explores the interactions between social biases and automatic topic detection models, and their impact on the resulting data collection. Our goal is to help social scientists gain insights about biases in text analytic so as to mitigate such biases in their data collections.

More specifically, we examine the role of political ideology biases (liberal-leaning, denoted as {\em Blue}, and conservative-leaning, denoted as {\em Red}) in the process of collecting data about certain social topics ({\em immigration} and {\em gun control}) from Twitter.  
We observe that biases may be introduced at three major junctures of the data collection pipeline. First, it may be introduced in the data source itself (\emph{e.g.}, certain forums may have strong political leanings), but social scientists typically choose their data intentionally and are aware of pre-existing biases therein \cite{malik2015population,kosinski2015facebook,cihon2016biased}. Second, biases may be introduced in the way in which \say{topic relevance} is defined. For example, domain experts may be consulted to identify a set of keywords or sample instances that are indicative or representative of the topic of interest.
Thus, any unconscious bias on the part of the domain experts would be encoded into these keywords and examples  \cite{king2017computer}, which would then serve as a noisy training corpus for developing a topic classifier. Third, the choice of the computational models for performing relevance classification may amplify or mitigate the impact of the biases. 

Through a suite of empirical analyses, this work studies the effect of biased keywords
(\textit{Blue}-leaning, \textit{Red}-leaning) on downstream training and retrieval: 1) To what extent does a trained classifier propagate the bias seen in the training data? Can it learn to generalize and blunt some of the bias? 2) To what extent do biases in the training corpus degrade the overall retrieval ability of the classifier? Specifically, we
generate strongly \textit{Blue}-leaning and \textit{Red}-leaning noisy training sets, and we compare the impact of these training sets on 
three common off-the-shelf models: GloVe \cite{pennington2014glove}, ELMo \cite{peters2018deep}, and BERT \cite{devlin2018bert}. The three models are chosen to span a range of model sizes and representational power. 
We find that of the three off-the-shelf models, BERT more frequently suffers a significant drop in retrieval quality and propagates more bias when trained on biased data.


We then propose a method to mitigate the bias. That is, we want a classifier that is oblivious to an instance's group affiliation to \textit{Blue} or \textit{Red}, yet still performs the main task of judging the instance's 
relevance to the topic. Our approach adapts Domain-Adversarial Training \cite{ganin2016domain} for the three off-the-shelf models.  
Experimental results show that the proposed approach mitigates the unintended bias at no or little cost of retrieval accuracy as compared to the original models; in fact, the retrieval accuracy for the modified BERT is slightly boosted. The code and data for this project is  available.\footnote{https://github.com/MeiqiGuo/COLING2020-BiasStudy}

\section{Political Ideology Bias on Social Topic Detection}
\label{sec:political_bias}
We investigate the impact of political ideology biases on extracting tweets relevant to specific social topics. Unlike gender or racial bias, which has been widely studied in language representation, machine translation or relation extraction \cite{stanovsky-etal-2019-evaluating,gaut-etal-2020-towards,blodgett2017racial}, there exists fewer work on political ideology biases.
Political ideology biases on social topic detection may arise from the difference in language usages between political ideological groups. 
Prior studies have compared different language usages between political ideological groups such as conservatives and liberals in the US. Such differences are reflected in general linguistics patterns such as language complexity \cite{schoonvelde2019liberals} and emotions associated with language \cite{wojcik2015conservatives}. Moreover, while talking about the same topic, language devices, such as specific types of metaphors, are often found to be different, which are associated with the groups’ distinct political background and moral concerns \cite{dehghani2011analyzing,lakoff1995metaphor}. These observations indicate that  the information producers come from diverse political ideological backgrounds, and the selection of keywords is critical for obtaining balanced and representative data points. 
Therefore, we first examine the political ideology biases introduced from in human-selected keywords.

\subsection{Data Source: Twitter}
\label{sec:data}
We focus on Twitter because it is a widely-used space for people to express their views on social topics. For our study, we rely on a prior work \cite{yang2017quantifying} that collected data from publicly posted tweets using official Twitter APIs during a time-frame close to the 2016 U.S. Presidential election. 
Two groups of users are identified -- Clinton-supporters (\textit{Blue}) and Trump-supporters (\textit{Red}) -- that are likely to have distinct political and ideological preferences. Group membership is defined as an  \textit{exclusive follower}; \emph{i.e.}, Twitter users who followed only one presidential election candidate but not the other. In their study, Yang et al. \shortcite{yang2017quantifying} have validated the concept that exclusive followers make good proxies for group affiliations. 
Our final raw corpus used for this study consists of over 7 million tweets. 
More detail for our data collection is described in Section \ref{sec:setting}. 

\subsection{Quantifying Bias in Keywords}
\label{subsec:keyword_bias}
For any controversial topic, some useful keywords are necessarily going to be biased toward one group or another. Even taken as a set, the keywords that a human expert came up with may reflect the bias of that expert. A useful piece of information, therefore, is if we could quantify the level of bias in the keywords. It could inform the experimenters on whether they should recruit additional diverse experts to expand their keyword set.

Since we have the ground truth for the political group (\textit{Red}/\textit{Blue}) of each tweet, we could use a simple ratio between the 
numbers of tweets containing keyword $x$ between one group with the other as the metric, but a better metric is to apply a Chi-square test because 
it takes deviation into consideration for estimating the probability.
More specifically, we evaluate the bias of each keyword by two-tailed Pearson Chi-square Test (details in Appendix \ref{app:chi-square}). The root challenge of this quantifying methodology is that due to the sheer size of the raw corpus, we do not have the full ground truth for topic relevance (\textit{i.e.}, whether a tweet is about gun control or immigration for all tweets). 
A direct consequence is that we cannot identify all the high-precision keywords by brute-force; thus, our study also relies on human-chosen keywords. Even though keywords chosen by an expert may be biased, the ensemble of keywords from diverse experts is much less likely to be  biased \cite{king2017computer}. Therefore, we approximate the ground truth for topic relevance by the ensemble of human-chosen keywords for the Chi-square test of each single keyword.

We selected the topics of gun control and immigration from the ProCon website\footnote{This website has organized and collected major arguments and researches relevant to controversial issues in the US, in which the information was arranged into pros and cons that reflect the opposite stances and ideas around the issues.} because both have engaged enthusiastic political debates, with  extremely  conflicting  stances  and opinions from opposing political camps. Keywords are collected from diverse experts who are familiar with or have worked on these social topics in tweet corpus. There are 29 keywords for the immigration topic and 34 keywords for the gun control topic. We assign each keyword to the \textit{Blue}-leaning, \textit{Red}-leaning or unbiased (neutral) group by setting the confidence level of the Chi-square test equal to 99\%. Table \ref{tab:keyword} shows the number of keywords in each group as well as some examples, which reveals that most (around 75\%) expert-selected keywords actually have political ideology bias. Moreover, some keywords are extremely biased, such as \say{\#NoBanNoWall} for the topic immigration and \say{\#NoBillNoBreak} for the topic gun control (refer to Appendix \ref{app:zscore} which shows the exact Z-test scores for each keyword).
Our findings verify our hypothesis that the language usages by different political ideology groups are often found to be different, even while talking about the same topic. These observations suggest that a perfectly balanced selection of keywords or a fully representative set of data points of diverse political ideological camps may not be achievable in practice. Therefore, there is a pressing need to study how biases propagate through topic detection models when they are trained on biased keywords.
\begin{table}
\centering
\begin{tabular}{p{2cm} p{1cm} p{5cm} p{1cm} p{4cm}}
\cline{2-5}
&\multicolumn{2}{l}{\textbf{Immigration}}&\multicolumn{2}{l}{\textbf{Gun control}}\\
\cline{1-5}
Bias Group&Num & Example&Num & Example\\
\hline
Blue&9&\#NoBanNoWall, \newline undocumented immigrant&12&\#NoBillNoBreak,\newline gun reform\\\hline
Neutral&6&deportation, \newline asylum&8&gun regulation,\newline \#momsdemandaction\\\hline
Red&14&illegal alien, \newline\#buildthewall&14&second amendment, \newline gun free\\\hline
\end{tabular}
\caption{Number and examples of keywords in each bias groups.}\label{tab:keyword}
\end{table}
\section{Bias Propagation through Models}
\label{sec:model_bias}
Unlike well-curated and annotated benchmark datasets, raw social media data is sprawling and unorganized.
Contributors come from diverse backgrounds, with different racial origins, personalities, education levels, etc.; they may hold many kinds of implicit biases, some of which may not have been identified by the social scientists carrying out the experiment. Under this setting, prior work for addressing the bias and ethical issues such as data statements \cite{bender2018data} may not be applicable. Nonetheless, data sources such as Twitter remain a powerful resource that researchers are willing to tap into. Therefore, it is important to compare how different NLP systems perform on potentially biased training data and to develop approaches for mitigating bias propagation through models. 

We consider bias propagation in two dimensions: 1) To what extent does a model trained on biased examples tend to detect more instances with the same bias? 2) How does the learned bias interact with {\em relevance}? (Does a biased classifier simply retrieve fewer instances of the other group, or does it actually retrieve less relevant instances for that group?). We also want to determine whether certain types of NLP systems are more likely to propagate the bias. Given that NLP models are built with a diverse of architectures (transformers, RNN, etc.) and the number of trainable parameters varies from hundreds to billions, we define the type of NLP systems along their context representations and sizes. 

Prior work shows that complex models, such as BERT, do quite well for many NLP applications. Multi-head attention allows BERT to be able to capture complex and fine-grained patterns for the target prediction. On the other hand, big complex models with numerous training parameters are more likely to be overfitted when there are not enough training data \cite{yin2018dimensionality}. Therefore, it is not obvious what might happen with biased-trained large complex models: do they succeed in using "real" patterns for the target task (\textit{e.g.}, predict relevant tweets in our case); or do they make use of the bias seen in the training data (\textit{e.g.}, capitalize on superficial patterns in the biased data) for reaching a minimum loss? Our work aims to answer this question by examining three representative NLP models under multiple, differently biased training sets. 

\subsection{Comparison between Different Off-the-shelf NLP Models}
\label{sec:compare_model}

Our study is over two state-of-the-art NLP models, representing high performance approaches, and one simpler model, representing the benchmark. We compare three different topic detection models which are respectively built with BERT, ELMo, and GloVe. For predicting relevant tweets of a target topic, we fine-tune the BERT model with just one additional output layer. When we build topic detection models using ELMo and GloVe, we add a Bi-LSTM layer after ELMo/GloVe as the text encoder, then feed it forward through one output layer for predicting relevance.  These three text encoder models have different architectures and sizes, shown as below.

\textbf{BERT} \cite{devlin2018bert} A language representation model whose architecture is deep bidirectional transformers and which is pre-trained on large-scale unlabeled text corpus. It can be fine-tuned with just one additional output layer to create state-of-the-art models for a wide range of tasks, such as question answering and language inference. The base BERT model has 110M trainable parameters. 

\textbf{ELMo} \cite{peters2018deep} A large-scale pre-trained deep contextualized word representation. Contextual word vectors are learned functions of the internal states of a deep bidirectional language model. These representations significantly improve the state of the art across six challenging NLP problems, including question answering, textual entailment and sentiment analysis, at the time it was released. Our topic detection model built with ELMo has 3M trainable parameters.

\textbf{GloVe} \cite{pennington2014glove} A traditional distributed word representation learnt from global log-bilinear regression model of word-word co-occurrence matrix. It could capture fine-grained semantic and syntactic regularities using vector arithmetic. Our topic detection model built with GloVe has 16M trainable parameters (the number of parameters is linearly correlated with the vocabulary size).

We intentionally experiment with these three NLP models in order to answer the question about the relation between model complexity and robustness to bias, because they are respectively good representatives of bidirectional transformers, bidirectional LSTM and single word vectors. Moreover, their parameter sizes are in different magnitudes.

\subsection{Proposed Approach for Mitigating Bias Propagation}

One promising explanation for bias propagation of ML models is that inductive bias in gradient descent methods results in the overestimation of the importance of moderately-predictive “weak” features if training data is biased and insufficient \cite{jayakumar2019multiplicative}. Due to the difference in language use between political ideological groups, topic detection models are biased towards learning frequent spurious correlations in the training data instead of learning true indicators of relevance. For example, when most immigration-relevant tweets are posted by users in \textit{Blue} while \textit{Blue} and \textit{Red} users are evenly distributed in non-relevant tweets in the training data, the text classification systems may overestimate the importance of \textit{Blue} users language features as the signal of relevance. This could result in a lose of retrieval accuracy for both \textit{Blue} and \textit{Red} tweets and especially a low recall for \textit{Red} tweets during the test time. 

The reason for bias propagation of ML models is close to domain adaptation, in which training and test data come from similar but different distributions. Ben-David \shortcite{ben2007analysis} suggests that a good representation for cross-domain transfer is one for which an algorithm cannot learn to identify the domain of origin of the input observation. We expect something similar here: an ideal representation of tweets should be invariant to group affiliation (\textit{Blue}/\textit{Red}) as well as discriminant to topic relevance. With this thought, we propose an approach inspired by a domain adaptation technique -- Domain-Adversarial Neural Networks \cite{ganin2016domain} as a way to mitigate the bias propagation. Prior work uses adversarial feature learning for demoting latent confounds with respect to the task of native language identification \cite{kumar2019topics}; for interpreting computational social science with deconfounded lexicon induction \cite{pryzant2018deconfounded}; or for preserving privacy by removing demographic attributes \cite{li2018towards}. Xie \shortcite{xie2017controllable} demonstrates the effectiveness of adversarial feature learning on fair classifications. However the datasets they used (for predicting the savings, credit ratings and health conditions of individuals) have no natural language text input. Our work focuses on state-of-the-art NLP models (such as BERT) and applies adversarial feature learning to mitigate the political ideology bias.

\begin{figure}
    \centering
    \includegraphics[scale=0.7]{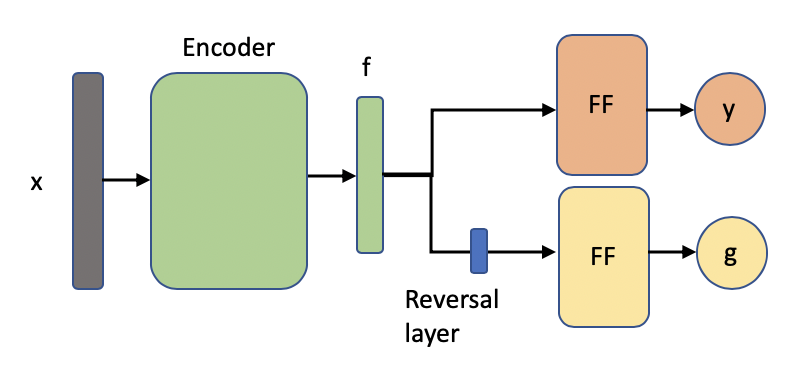}
    \caption{Model architecture of our proposed approach. Encoder could be BERT, ELMo+BiLSTM or GloVe+BiLSTM. The top Feed Forward NN (in orange) is the class label predictor. The bottom Feed Forward NN (in yellow) is the political group predictor.}
    \label{fig:model}
\end{figure}

Our goal is to train a classifier that learns to accurately predict the relevance, while ignoring superficial patterns biased with political group present in the training set. The architecture of our proposed model is shown in Figure \ref{fig:model}. The input tweet $x$ first goes through a text encoder $e(x;\theta_{e})$ for getting a feature vector $f_{x}$ as the text representation. The encoder could be BERT, ELMo+BiLSTM or GloVe+BiLSTM, exactly same as what we describe in Section \ref{sec:compare_model}. Then the feature vector $f_{x}$ is fed into two one-layer feed forward neural networks: 1) $r(f_{x};\theta_{r})$ (FF in orange in Figure \ref{fig:model}) for predicting whether $x$ is relevant or not; 2) $g(f_{x};\theta_{g})$ (FF in yellow in Figure \ref{fig:model}) for predicting whether $x$ is posted by a \textit{Blue} or \textit{Red} user. As gradients back-propagate from the group prediction $g$ heads to the encoder, we pass them through a gradient reversal layer \cite{ganin2016domain}, which multiplies gradients by $-1$. If the cumulative loss of the relevance prediction is $L_r$ and that of the group classification is $L_g$, then the loss which is implicitly used to train the encoder is $L_e = L_r-\alpha L_g$ (with loss weighted by $\alpha$), thereby encouraging the encoder to learn representations of the text which are not useful for predicting the political group. We use Cross-Entropy for computing $L_r$ and $L_g$: $$L_r=\frac{1}{N}\sum_{i=1}^{i=N}CE(r(e(x_i)),y_i)$$
$$L_g=\frac{1}{N}\sum_{i=1}^{i=N}CE(g(e(x_i)),g_i)$$

In this work we use those three off-the-shelf NLP models as text encoder for comparing directly with Section \ref{sec:compare_model}, but the proposed approach is applicable to other encoder models as well.


\section{Experiments}
\label{sec:experiments}

To address the central questions raised in this work -- how biases are propagated in several widely-used NLP models and their effect on the overall retrieval accuracy, we conduct experiments to quantify the impact of biased training. To do so, we need to generate training sets for which we can measure the degree of bias along the \textit{Blue-Red} spectrum. We also need metrics for determining the quality and bias of retrieved tweets. With this evaluation framework, we compare how different NLP models perform under multiple, differently biased training sets on two social topics -- immigration and gun control. 
Then we evaluate the effectiveness of our proposed bias mitigating approach under the same setting.

\subsection{Experimental Setup}
\label{sec:setting}
\textbf{Data:}
In this study, we build on data acquired from prior work by Yang et al.\shortcite{yang2017quantifying}. They collected over 7 million tweets posted by the \textit{exclusive followers} of Trump and Clinton within a nine-month period (between June 2016 and February 2017). 
We pre-process this Tweet corpus by removing emoji, website links and usernames. Then we split it into training and test set by a ratio of 9:1. Topic detection models are trained and validated on the training corpus, and the retrieval quality and retrieval bias are evaluated on the test set.

\textbf{Training Set Settings:}
For each topic we collect a set of keywords (referred as $K_{total}$) from experts and compute their bias scores (Section \ref{subsec:keyword_bias}). Keywords that don't pass the Chi-square test at the confidence level equal to 99\% are considered as biased. If the z-score of a biased keyword is positive, it is biased towards \textit{Blue}; otherwise, it is biased towards \textit{Red}. We refer the set of \textit{Blue}-leaning keywords  as $K_{blue}$ and \textit{Red}-leaning keywords as $K_{red}$. For each keyword in $K_{blue}$ (respectively $K_{red}$), we extract all tweets containing those keywords from the training corpus as \say{relevant} examples;  we randomly select an equal number of tweets that don't contain any keyword as \say{irrelevant} examples. In this way we construct a noisy training dataset biased towards \textit{Blue} (respectively \textit{Red}). Similarly, we also construct a full training dataset with $K_{total}$. We report the number of relevant tweets by this keyword approach in the training and test set in Table \ref{tab:size}. Notice that for the topic of immigration the size of \textit{Blue}-leaning training set is twice more than \textit{Red}-leaning training set.

\textbf{Model Settings:}
We use the publicly available versions of BERT (bert-base-uncased\footnote{https://storage.googleapis.com/bert\_models}), ELMo (orginal\footnote{https://allennlp.org/elmo}) and GloVe (common crawl 830B300d\footnote{https://nlp.stanford.edu/projects/glove/}) with the recommended parameter settings. The loss weight $\alpha$ for the adversarial training gradually increases from 0 to 1 along with the increasing of training batch number. When training our bias mitigating approach, we keep most hyper-parameters same as its correspondent original model, \textit{e.g.}, learning rate, epoch number, batch size, maximum sequence length, etc. For ELMo+ADV we tune the dropout parameter and report the one without RNN input dropout in Section \ref{sec:exp_new_approach}.

\begin{table}
\centering
\begin{tabular}{ll|l||l|l}
\cline{2-5}
&\multicolumn{2}{c||}{\textbf{Immigration}}&\multicolumn{2}{c}{\textbf{Gun control}}\\
\cline{2-5}
&Training & Test&Training & Test\\
\hline
\textbf{$K_{blue}$} & 17,755&2,061&4,340&486\\
\textbf{$K_{red}$} & 8,606&947&4,218&509\\
\textbf{$K_{total}$} & 26,642&3,007&9,244&1,049\\\hline
\end{tabular}
\caption{Number of relevant tweets by the keyword approach in the training and test set.}\label{tab:size}
\end{table}

\textbf{Evaluation Metrics:}
While our data contains the ground truth for whether an instance belongs to \emph{Blue} or \emph{Red}, we do not know a priori whether it is relevant to some topic or not. To determine model performance on the dimension of relevance, we use {\em precision at $N$}, which computes the precision score of a set of instances from the top $N$ prediction scores. The relevance is judged by crowdsourcing workers via Amazon Mechanical Turk. 
In order to evaluate the retrieval quality, we need to choose a reasonable $N$ for the metric \textit{P@N}: if $N$ is too large or too small, then all models would have a very low or high precision score so that the comparison between them is not significant.
We find that the number of relevant tweets extracted by $K_{total}$ from the test set is a good candidate for \textit{N}. It is not too large because we expect there to be at least this many relevant tweets in the test set; and it is not too small, since keywords $K_{red}$ or $K_{blue}$ which are used for generating training sets are subsets of $K_{total}$. As Table \ref{tab:size} shows, there are 3007 tweets and 1049 tweets which contain keywords from $K_{total}$ for immigration and gun control, respectively. Therefore, we use \textit{P@3000} and \textit{P@1000} (round 3007 and 1049 to hundred) as evaluation metric respectively for immigration and gun control. For reducing the annotation cost, we randomly select 100 samples from the top 3000 or 1000 for human annotation which could significantly represent the performance of models.
To determine the level of bias in a model's predictions, we compute the \emph{Blue}-versus-\emph{Red} {\em Log Odds Ratio} to determine how likely a retrieved instance (one of the top $N$) is from \textit{Blue} instead of the \textit{Red}. The first odds computes \# of \emph{Blue} instances: \# of \emph{Red} instances in top $N$; the second odds computes the same odds in the non-retrieved instances.

\textbf{Annotation Process:}
The ground-truth relevance of retrieved tweets is annotated by crowdsourcing workers via Amazon Mechanical Turk. Selected workers are well trained and carefully evaluated by qualification tests in order to make sure they know the coverage of each social topic (refer to Appendix \ref{app:annotation}), for example immigration covers a broad set of sub-topics ranging from a specific policy (\emph{e.g.}, DACA), border security, birthright citizens, to labor market. We add gold standard instances to each annotation batch for monitoring the annotation quality. In addition, each instance is annotated by two annotators and a third person is involved if they don't meet an agreement. The average accuracy and inter-agreement of our annotations are both above 90\%. 
\subsection{Results of Comparing Off-the-shelf NLP Models}
\label{sec:exp_compare_models}
\begin{table}
\centering
\begin{tabular}{llll|lll||lll|lll}
\cline{2-13}
&\multicolumn{6}{c||}{\textbf{Immigration \textit{P@3000}}}&\multicolumn{6}{c}{\textbf{Gun control \textit{P@1000}}}\\
\cline{2-13}
&\multicolumn{3}{c|}{\textit{Blue}-leaning Train}&
\multicolumn{3}{c||}{\textit{Red}-leaning Train} &
\multicolumn{3}{c|}{\textit{Blue}-leaning Train} &
\multicolumn{3}{c}{\textit{Red}-leaning Train} \\\cline{2-13}
&
All&
\textit{Blue}&
\textit{Red}&
All&
\textit{Blue}&
\textit{Red}&
All&
\textit{Blue}&
\textit{Red}&
All&
\textit{Blue}&
\textit{Red}\\
\hline
Baseline&
0.66&
0.66&
0.66&
0.29&
0.30&
0.28&
0.44&
0.44&
0.36&
0.46&
0.46&
0.44\\\hline
GloVe&
\color{red}{0.77}&
0.72&
0.89&
0.68&
0.70&
0.65&
0.74&
0.81&
0.59&
\color{red}{0.66}&
0.64&
0.68\\
ELMo&
\textbf{0.86}&
0.85&
0.88&
\textbf{0.69}&
0.67&
0.71&
0.77&
0.78&
0.74&
\textbf{0.75}&
0.73&
0.77\\
BERT&
\color{red}{0.74}&
0.73&
0.78&
\textbf{0.69}&
0.67&
0.72&
\textbf{0.79}&
0.80&
0.75&
\color{red}{0.59}&
0.66&
0.52\\\hline
\end{tabular}

\caption{Retrieval accuracy of different topic detection models trained on \textit{Blue}-learning or \textit{Red}-leaning training sets. Columns \say{All}, \say{\textit{Blue}} and \say{\textit{Red}} respectively show the accuracy for all retrieved tweets, retrieved tweets posted by \textit{Blue} users and retrieved tweets posted by \textit{Red} users. Within each column of \say{All}, the best model is bolded; if a model's performance is over 10\% worse than the best one, then it is marked in red color.}
\label{tab:result}
\end{table}

The retrieval quality of different topic detection models trained on biased training sets is shown in Table \ref{tab:result}. Models built with GloVe, ELMo and BERT are trained on \textit{Blue}-leaning and \textit{Red}-leaning training sets. In addition to the three off-the-shelf models, we also include a naive keyword approach as baseline which only retrieves tweets containing training keywords. Let's first look at columns \say{All} which show the accuracy for all retrieved tweets. For the both topics, it is not surprising that trained NLP models generally outperform the keyword-extraction baseline. This means that the models are able to learn some patterns besides the keywords and generalize to tweets which do not contain any keyword. For more easily comparing between models, the best model within the column is bolded; if a model's performance is over 10\% worse than the best one, then it is marked in red color. Our experimental results show that ELMo-based model has the best overall retrieval quality; BERT-based model is the most negatively affected by the training bias. Next, we compare models' accuracy for retrieved tweets posted by \textit{Blue} users and \textit{Red} users by looking into columns \say{\textit{Blue}} and \say{\textit{Red}}. In general, models have a better retrieval accuracy for tweets from the group towards which the training set is biased (as the second, third and fourth big column show), except for the first big column (when models are trained on the \textit{Blue}-leaning set for the topic immigration, models have a better retrieval accuracy for \textit{Red} tweets than \textit{Blue}). 
We also report the retrieval accuracy for models trained on the full training set (constructed from $K_{total}$) in Table \ref{tab:result_total}. The performance of different models are close to each other.

\begin{table}
\centering
\begin{tabular}{cc||c}
\cline{2-3}
&\textbf{Immigration \textit{P@3000}}&\textbf{Gun control \textit{P@1000}}\\
\hline
Baseline&
0.86&
0.83
\\
GloVe&
0.88&
0.85
\\
ELMo&
0.93&
0.82
\\
BERT&
0.89&
0.83
\\\hline
\end{tabular}
\caption{Retrieval accuracy for models trained on the full training set (constructed from $K_{total}$).}
\label{tab:result_total}
\end{table}

Next, we evaluate to what extent the political bias is propagated to the retrieved (predicted top $N$) tweets by different NLP models. \emph{Blue}-versus-\emph{Red} \textit{Log Odds Ratio} of different topic detection models trained on \textit{Blue}-leaning or \textit{Red}-leaning training sets are shown in Table \ref{tab:ratio}. We use the bias in each training set as baseline. The closer to 0 \textit{Log Odds Ratio} is, the less the political bias is propagated. Positive means leaning to \textit{Blue} and negative means leaning to \textit{Red}. Experimental results show that for both topics, NLP models are all able to mitigate political ideology bias from training data. Especially for models trained on \textit{Blue}-leaning set of the topic gun control, the initial training set is highly biased towards \textit{Blue} group with an \textit{Log Odds Ratio} equal to 2.08, NLP models are able to mitigate 61\% less bias. For comparing between models more easily, the best model within each column is bolded. We find that ELMo and GloVe-based models propagate the least of the bias; BERT-based model propagates the most of the bias seen in the training data.

\begin{table}
\begin{tabular}{cc|c||c|c}
\cline{2-5}
&\multicolumn{2}{c||}{\textbf{Immigration}}&\multicolumn{2}{c}{\textbf{Gun control}}\\\cline{2-5}
&\textit{Blue}-leaning Train&\textit{Red}-leaning Train & \textit{Blue}-leaning Train & \textit{Red}-leaning Train\\
\hline

Training set&
0.81&
-0.16&
2.08&
-0.29\\\hline
GloVe&
0.69&
\textbf{0.10}&
\textbf{0.82}&
-0.08\\
ELMo&
\textbf{0.60}&
0.14&
0.84&
\textbf{-0.02}\\
BERT&
0.62&
0.14&
0.94&
0.05\\\hline
\end{tabular}

\caption{\emph{Blue}-versus-\emph{Red} \textit{Log Odds Ratio} of different topic detection models trained on \textit{Blue}-leaning or \textit{Red}-leaning training sets. Within each column, the best model is bolded. The closer to 0, the better (less bias). Positive means leaning to \textit{Blue}; negative means leaning to \textit{Red}.}
\label{tab:ratio}
\end{table}

Taking both the retrieval accuracy and retrieval bias into consideration, we conclude that ELMo-based model is the most robust to training bias, while BERT-based model is the most negatively affected by the training bias. Our findings inform practitioners to choose the more robust model when training data is biased. Moreover, it is important to develop new approaches for mitigating the impact of bias, especially for BERT-based models. 

\subsection{Results of the Bias Mitigating Approach}
\label{sec:exp_new_approach}
\begin{table}
\centering
\begin{tabular}{p{2cm}p{1.5cm}|p{1.5cm}|p{1.5cm}||p{1.5cm}|p{1.5cm}|p{1.5cm}}
\cline{2-7}
&\multicolumn{3}{c||}{\textbf{Immigration \textit{P@3000}}}&\multicolumn{3}{c}{\textbf{Gun control \textit{P@1000}}}\\\cline{2-7}
& Full &\textit{Blue}-leaning&\textit{Red}-leaning & Full & \textit{Blue}-leaning & \textit{Red}-leaning\\
\hline

GloVe&
0.88&
0.77&
0.68&
0.85&
0.74&
0.66
\\
GloVe+ADV&
0.86&
0.72&
0.63&
0.73&
0.31&
0.65
\\\hline
ELMo&
0.93&
0.86&
0.69&
0.82&
0.77&
0.75
\\
ELMo+ADV&
0.82&
0.84&
0.59&
0.73&
0.66&
0.71
\\\hline
BERT&
0.89&
0.74&
0.69&
0.83&
0.79&
0.59
\\
BERT+ADV&
\textbf{0.90}&
\textbf{0.86}&
0.65&
0.79&
\textbf{0.87}&
\textbf{0.71}
\\\hline
\end{tabular}

\caption{Retrieval accuracy of original and our proposed models trained on full, \textit{Blue}-learning or \textit{Red}-leaning training sets. Our proposed model is bolded if it outperforms its base model.}
\label{tab:result_new}
\end{table}
We compare our bias mitigating approach with its original models on both retrieval accuracy and retrieval bias metrics. We report the performance of our bias mitigating approach in both a more realistic training scenario -- models are trained on the full training data, and extremely biased cases -- models are trained on our generated strongly \emph{Blue}-leaning or \emph{Red}-leaning training sets. The full training dataset is slightly biased towards \emph{Blue} with a \emph{Blue}-versus-\emph{Red} \textit{Log Odds Ratio} equal to 0.51 for the topic of immigration or 0.61 for the topic of gun control. Table \ref{tab:result_new} shows that our proposed BERT-based model (noted as BERT+ADV) improves the retrieval accuracy compared with the original one; for ELMo+ADV and GloVe+ADV the retrieval accuracy slightly reduces for most cases. Table \ref{tab:ratio_new} shows that our proposed models are very efficient for mitigating the bias propagation. In sum, experimental results demonstrate that our proposed approach succeeds to mitigate bias propagation, with no or little drop of retrieval accuracy. It works especially well for BERT-based models -- mitigates the bias at the same time increases the retrieval accuracy.

\begin{table}
\centering

\begin{tabular}{p{2cm}p{1.5cm}|p{1.5cm}|p{1.5cm}||p{1.5cm}|p{1.5cm}|p{1.5cm}}
\cline{2-7}

&\multicolumn{3}{c||}{\textbf{Immigration}}&\multicolumn{3}{c}{\textbf{Gun control}}\\\cline{2-7}
& Full &\textit{Blue}-leaning&\textit{Red}-leaning & Full & \textit{Blue}-leaning & \textit{Red}-leaning\\

\hline
GloVe&
0.43&
0.69&
0.10&
0.58&
0.82&
-0.08\\
GloVe+ADV&
0.43&
\textbf{0.54}&
\textbf{0.07}&
\textbf{0.14}&
\textbf{0.77}&
\textbf{0.01}\\\hline
ELMo&
0.45&
0.60&
0.14&
0.31&
0.84&
-0.02\\
ELMo+ADV&
\textbf{0.34}&
\textbf{0.52}&
\textbf{-0.12}&
\textbf{0.10}&
\textbf{0.35}&
\textbf{0.01}\\\hline
BERT&
0.40&
0.62&
0.14&
0.62&
0.94&
0.05\\
BERT+ADV&
0.47&
\textbf{0.57}&
\textbf{0.12}&
\textbf{0.23}&
\textbf{0.70}&
\textbf{0.00}\\
\hline
\end{tabular}

\caption{\emph{Blue}-versus-\emph{Red} \textit{Log Odds Ratio} of original and our proposed models trained on full, \textit{Blue}-leaning or \textit{Red}-leaning training sets. Our proposed model is bolded if it outperforms its base model. The closer to 0, the better (less bias). Positive means leaning to \textit{Blue}; negative means leaning to \textit{Red}.}
\label{tab:ratio_new}
\end{table}
\section{Conclusion}
We have studied the impact of political ideology biases in different types of topic detection models and demonstrated a domain adaptation approach as an effective way of mitigating the bias. Our experimental results suggest that an ELMo-based model is more robust to training bias, while a BERT-based model is more negatively affected by the training bias. Since the ELMo-based model has nearly 40 times fewer trainable parameters than BERT, we conjecture that big complex models are more likely to propagate the bias seen in the training set. Although we have found the proposed adaptation architecture to be helpful for the three models, especially BERT, in terms of mitigating some of the training bias, the approach still relies on some knowledge of the existence of the bias. 
This work offers a comparison point for future studies to evaluate the effect of bias in various predictive models and opens the door for further reducing the bias in topic detection applications.
 
\section*{Acknowledgements}
The authors would like to acknowledge the support from the DARPA UGB and AFOSR awards. Any opinions, findings, and conclusions or recommendations expressed in this material do not necessarily reflect the views of the funding sources.

\bibliographystyle{coling}
\bibliography{coling2020}

\begin{appendices}
\section{Chi-square Test for evaluating keyword bias}
\label{app:chi-square}
Assume \emph{T} is the corpus of tweets, \emph{K} is the keyword set, the bias dimension is towards either group \emph{$G_1$} or \emph{$G_2$}. $\forall x \in K$, the null hypothesis $H_0$ is that the probability of a keyword \emph{x} appears in relevant tweets posted by people from group \emph{$G_1$} and group \emph{$G_2$} is the same. For our problem, the keyword $x$ is unbiased if it passes the Chi-square Test with a specific confidence level; otherwise, $x$ is biased. The bias level is measured by Z-statistic. The statistic test formula is shown as below:
$$z_x=\frac{\hat{p_{1x}}-\hat{p_{2x}}}{\hat{\sigma_{Dx}}}$$
$$\hat{\sigma_{Dx}}=\sqrt{\hat{p_x}(1-\hat{p_x})(\frac{1}{n_1} + \frac{1}{n_2})}$$
$$\hat{p_x} = \frac{n_1\hat{p_{1x}}+n_2\hat{p_{2x}}}{n_1+n_2}$$
$$\hat{p_{1x}} = \frac{n_{1x}}{n_1}$$
$$\hat{p_{2x}} = \frac{n_{2x}}{n_2}$$
where $n_{i}$ is the number of relevant tweets posted by group \emph{$G_i$}, and $n_{ix}$ is the number of tweets containing keyword $x$ and posted by group \emph{$G_i$}, $\forall i \in \{1,2\}$. Here a tweet is considered as relevant if it contains at least one keyword from $K$.

A cut-off score $z_{cut}$ is computed by the confidence level based on Z-test distribution. If $|z_x| < z_{cut}$, then the keyword $x$ is considered as unbiased; if $z_x > z_{cut}$, then the keyword $x$ is considered as biased towards group \emph{$G_1$}; if $z_x < -z_{cut}$, then the keyword $x$ is considered as biased towards group \emph{$G_2$}. In this way, keywords in the set $K$ could be divided into three groups: \emph{$G_1$}-leaning, \emph{$G_2$}-leaning and neutral. Additionally, more is the $|z_x|$, more biased is the keyword $x$.

\newpage
\section{Political ideology bias of each keyword}
\label{app:zscore}
The Z-test scores for each keyword are respectively shown in Table \ref{tab:keyword_bias} for immigration and gun control.
\begin{table}[h]
\centering
\begin{tabular}{p{5.5cm} p{1.5cm} p{5.5cm} p{1.5cm}}
\hline
\multicolumn{2}{l}{\textbf{Immigration}}&\multicolumn{2}{l}{\textbf{Gun control}}\\
\hline
Keyword & Z score & Keyword & Z score\\
\hline
nobannowall & 25.83 & NoBillNoBreak&25.05\\
muslimban&20.10 & gunviolence&14.32\\
refugee & 10.37 & disarmhate&13.53\\
immigration ban & 5.14 & gun violence&13.07\\
muslim ban & 4.66 & endgunviolence&9.64\\
immigration order & 3.89 & gunsense&8.10\\
immigrant & 3.36 & guncontrol&6.85\\
travel ban & 2.99 & WearOrange&6.21\\
undocumented immigrant & 2.61 & gun reform&6.17\\
&&gun safety&4.86\\
&&momsdemand&4.01\\
&&GunControlNow&2.84\\\hline
deportation & 1.10 &PrayForOrlando&1.79\\
executive order & 0.33 &OrlandoShooting&1.77\\
undocumented immigration & 0.22&NationalGunViolenceAwarenessDay&1.10\\
build a wall & -0.56 & momsdemandaction&1.10\\
asylum & -1.10 &gun regulation&0.90\\
deporting & -2.05 &gun death&0.87\\
&&gunskillpeople&-0.11\\
&&progun&-1.98\\\hline
undocumented alien & -2.88&gunright&-4.09\\
mexican wall & -3.10&gun law&-4.44\\
unvetted refugees & -0.48&SecondAmendment&-4.65\\
bansyrianrefugees & -0.48&gun right&-6.38\\
sanctuary city & -5.36&molonlabe&-6.43\\
amnesty & -7.29&gungrab&-6.60\\
immigration & -8.67&second amendment&-6.86\\
deported & -11.04&2ndamendment&-10.58\\
deport & -13.52&gun owner&-12.15\\
build the wall & -14.31&gun control&-13.09\\
illegal immigration & -16.38&gun free zone&-14.73\\
buildthewall & -18.54&2nd amendment&-14.82\\
illegal immigrant & -22.21 &gun free&-15.79\\
illegal alien & -23.29&firearm&-16.45\\\hline
\end{tabular}
\caption{Keywords and their $z$ score for immigration and gun control. Keywords at the top are blue-leaning; those in the middle are neutral; those at the bottom are red-leaning.}\label{tab:keyword_bias}
\end{table}

\section{Annotation Guidelines}
\label{app:annotation}
\subsection{Instructions for annotating the topic of immigration}
A tweet is considered as relevant if it talks about anything that \textbf{has to do with, but not limited to}, the following issue categories: Borders, Birthright citizenship, Immigrant Crime, DACA and the DREAM Act, Deportation debate, Economic impact, Immigration quotas, Immigrants’ rights and access to services, Labor Market - American workers and employers, Law enforcement, Muslin Ban/Travel Ban, Obama Iraq ban, Refugees, etc.

An example for the \textbf{option 1 - Relevant}: \say{Wonder what advice she got regarding her open border plan and especially her willingness to increase Syrian immigration?} This tweet talks about the open border plan and Syrian immigration, which is related to the topic of immigration under the categories of Border and Refugees; 

An example for the \textbf{option 2 - Not Relevant}: \say{'Will I die, miss?' Terrified Syrian boy suffers suspected gas attack.} This tweet talks about a Syrian boy suffering a gas attack, which may be pointing to a war or terrorist event in Syria, not necessarily directly about an immigration issue. 

\textbf{Instruction of some cases that may be more ambiguous}: 

A tweet should be considered as \textbf{relevant} if it: 1) mentions several topics in addition to immigration: \say{I'm a woman that supports Trump to fix economy, immigration, school, military \& more. \#MAGA3X}; 2) is short with relevant hashtags: \say{Against! \#muslimban}; 3) talks about immigration in other countries: \say{The \#EU referendum has become a sinister attack on immigrants \#Brexit \#Xenophobia.};

A tweet should be considered as \textbf{irrelevant} if it mentions a group of immigrant people such as Muslim, Syrian refugees but doesn't explicitly talk about immigration issues: \say{Wonderful news, I will suffer being video taped shopping with my wife, while Muslim terrorists construct bombs}, or \say{Syrian girl, 7, who tweeted from Aleppo meets Turkey's Erdogan by \#Reuters.}

\subsection{Instructions for annotating the topic of gun control}
A tweet is considered as relevant if it talks about anything that \textbf{has to do with, but not limited to}, the following issue categories: the Second Amendment, Gun control laws, etc. Tweets which contain the following hashtags are probably relevant to gun control: \#NoBillNoBreak, \#WearOrange, \#EndGunViolence, \#DisarmHate, \#molonlabe, etc. 

Some examples for the \textbf{option 1 - Relevant}: 1) \say{Standing up for the second amendment and carrying a firearm for self defense.} This tweet talks about the 2nd amendment, which is related to the topic of gun control; 2) \say{I don't understand why we can't ban assault weapons. We all know they are only used for hunting people. \#PrayForOrlando \#guncontrolplease.} This tweet talks about banning weapons and contains the hashtag \say{\#guncontrolplease}, which is relevant to the topic of gun control; 3)  \say{Stay Strong , Represent the American people \#NoBillNoBreak \#DisarmHate.} This tweet contains hashtags \say{\#NoBillNoBreak} and \say{\#DisarmHate} which are both relevant to gun control issues.

Some examples for the \textbf{option 2 - Not Relevant}: 1) \say{Apple replaced the gun emoji with a water gun in iOS 10.} This tweet talks about gun emoji, which is not related to a gun control issue; 2) \say{GUN GAME in MWR! WINTER CRASH EDITION!} This tweet talks about gun games, which is not related to a gun control issue; 3) \say{The Fourth Amendment protects you from unreasonable searches and seizures.} This tweet talks about the fourth amendment which is irrelevant to gun control issues. 

\textbf{Instruction of some cases that may be more ambiguous}: 

A tweet should be considered as \textbf{relevant} if it: 1) mentions several topics in addition to gun control: \say{I'm a woman that supports Trump to fix economy, immigration, school, gun control \& more.}; 2) is short with relevant hashtags: \say{This is good. \#NoBillNoBreak.};

A tweet should be considered as \textbf{irrelevant} if it mentions a gun death event or a gun violence news, but the context is not about gun control: \say{Love will always conquer hate. \#PrayForOrlando \#OrlandoShooting}; \say{Same ppl who just yesterday were praying against \#LGBTQ ppl are now praying for the \#LGBT victims. \#Hypocrisy \#PrayForOrlando \#p2 \#Orlando}; \say{He unknowingly followed us to LA. After he raped a kid \& killed a dude in OR the LA 5-0 caught him jacking a car. Gun shots were exchanged.}, or \say{Turkey car bomb and gun attack on courthouse in Izmir}.

\end{appendices}

\end{document}